\def\year{2022}\relax
\documentclass[letterpaper]{article} 
\usepackage{aaai22}  
\usepackage{times}  
\usepackage{helvet}  
\usepackage{courier}  
\usepackage[hyphens]{url}  
\usepackage{graphicx} 
\usepackage{natbib}  
\usepackage{caption} 
\DeclareCaptionStyle{ruled}{labelfont=normalfont,labelsep=colon,strut=off} 
\usepackage{algorithm}
\usepackage{algorithmic}
\usepackage{xspace}

\usepackage{newfloat}
\usepackage{listings}
\floatstyle{ruled}
\newfloat{listing}{tb}{lst}{}
\floatname{listing}{Listing}
\usepackage{amsmath,amsfonts,amssymb,bm}
\usepackage{pifont} 
\usepackage{graphicx,subfigure,epsfig,fancybox} 
\usepackage{float}
\usepackage{color} 
\usepackage{multirow}
\usepackage{natbib}

\usepackage{tikz}
\usetikzlibrary{shapes.geometric, positioning, calc}
\usetikzlibrary{arrows,shapes,calc}
\usetikzlibrary{trees,positioning,mindmap,shadows,fit}
\usetikzlibrary{decorations.pathreplacing}
\usetikzlibrary{tikzmark} 
\usetikzlibrary{intersections} 

\newcommand{\revised}[1]{\textcolor{blue}{}}
\renewcommand{\vec}[1]{\boldsymbol{#1}}

\usepackage{adjustbox}
\usepackage{array}
\usepackage{booktabs}

\newcolumntype{R}[2]{%
    >{\adjustbox{angle=#1,lap=\width-(#2)}\bgroup}%
    l%
    <{\egroup}%
}

\usepackage{setspace}

\usepackage[]{trackchanges}
\addeditor{YJ}

\newcommand{\ourmethod}{\textsc{FLAT}\xspace}
\usepackage{array}
\newcolumntype{P}[1]{>{\centering\arraybackslash}p{#1}}

\title{Adversarial Training for Improving Model Robustness? Look at Both Prediction and Interpretation}
\author {
   Hanjie Chen,
   Yangfeng Ji
}
\affiliations{
    Department of Computer Science \\
	University of Virginia \\
    \{hc9mx, yangfeng\}@virginia.edu
}

\begin{document}

\maketitle

\begin{abstract}
  Neural language models show vulnerability to adversarial examples which are semantically similar to their original counterparts with a few words replaced by their synonyms. 
A common way to improve model robustness is adversarial training which follows two steps—collecting adversarial examples by attacking a target model, and fine-tuning the model on the augmented dataset with these adversarial examples.
The objective of traditional adversarial training is to make a model produce the same correct predictions on an original/adversarial example pair. 
However, the consistency between model decision-makings on two similar texts is ignored.
We argue that a robust model should behave consistently on original/adversarial example pairs, that is making the same predictions (\textbf{\textit{what}}) based on the same reasons (\textbf{\textit{how}}) which can be reflected by consistent interpretations. 
In this work, we propose a novel feature-level adversarial training method named \ourmethod. 
\ourmethod aims at improving model robustness in terms of both predictions and interpretations.
\ourmethod incorporates variational word masks in neural networks to learn global word importance and play as a bottleneck teaching the model to make predictions based on important words. 
\ourmethod explicitly shoots at the vulnerability problem caused by the mismatch between model understandings on the replaced words and their synonyms in original/adversarial example pairs by regularizing the corresponding global word importance scores. 
Experiments show the effectiveness of \ourmethod in improving the robustness with respect to both predictions and interpretations of four neural network models (LSTM, CNN, BERT, and DeBERTa) to two adversarial attacks on four text classification tasks. 
The models trained via \ourmethod also show better robustness than baseline models on unforeseen adversarial examples across different attacks. \footnote{Code for this paper is available at \url{https://github.com/UVa-NLP/FLAT}}
\end{abstract}

\section{Introduction}
\label{sec:intro}

Neural language models are vulnerable to adversarial examples generated by adding small perturbations to input texts \citep{liang2017deep, samanta2017towards, alzantot2018generating}. 
Adversarial examples can be crafted in several ways, such as character typos \citep{gao2018black, li2018textbugger}, word substitutions \citep{alzantot2018generating, ren2019generating, jin2020bert, garg-ramakrishnan-2020-bae}, sentence paraphrasing \citep{ribeiro-etal-2018-semantically, iyyer-etal-2018-adversarial}, and malicious triggers \citep{wallace2019universal}.
In this paper, we focus on word substitution-based attacks, as the generated adversarial examples largely maintain the original semantic meaning and lexical and grammatical correctness compared to other attacks \citep{zhang2020adversarial}.

\begin{figure}[t]
	\centering
	\includegraphics[width=0.47\textwidth]{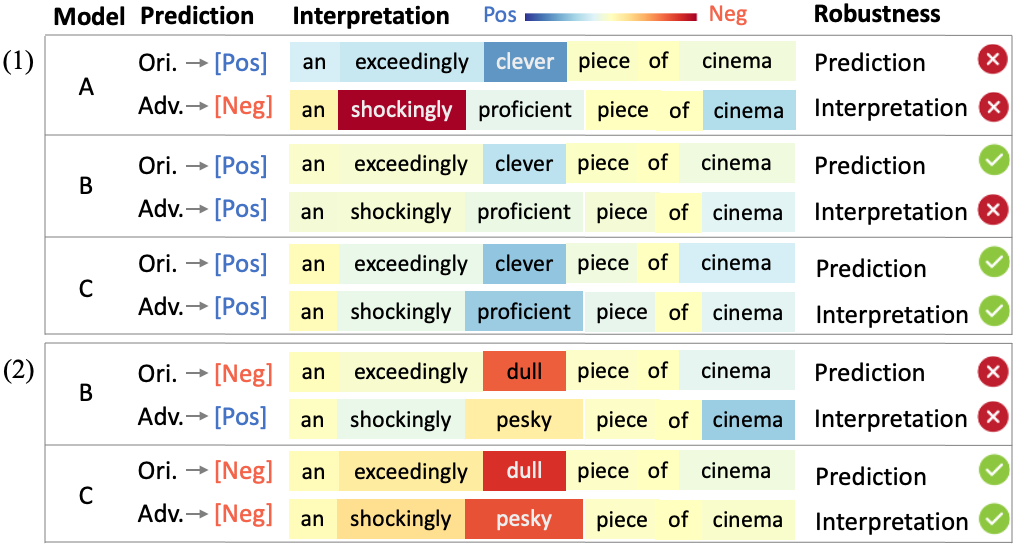}
	\caption{\label{fig:illustrations} Illustration of different model robustness with respect to predictions and interpretations on (1) a \textsc{positive} movie review and (2) a \textsc{negative} movie review (Ori.), and their adversarial counterparts (Adv.). Model B makes the same correct predictions on Ori. and Adv. in (1), while the discrepant interpretations reveal its vulnerability which is attacked by another adversarial example in (2). Only model C is robust with the same predictions and consistent interpretations on both original/adversarial example pairs.}
\end{figure}

Previous methods on defending this kind of attacks via adversary detection and prevention \citep{zhou-etal-2019-learning, mozes-etal-2021-frequency} or certifiably robust training \citep{jia2019certified, huang-etal-2019-achieving} either circumvent improving model predictions on adversarial examples or scale poorly to complex neural networks \citep{shi2020robustness}. 
Alternatively, adversarial training \citep{jin2020bert, li2020contextualized} improves model robustness via two steps—collecting adversarial examples by attacking a target model, and fine-tuning the model on the augmented dataset with these adversarial examples. 
However, existing adversarial training only focuses on making a model produce the same correct predictions on an original/adversarial example pair, while ignores the consistency between model decision-makings on the two similar texts.

To illustrate the necessity of maintaining consistent model decision-makings (reflected by interpretations) during adversarial training, Fig. \ref{fig:illustrations} shows both the predictions and their corresponding interpretations of different models on original/adversarial example pairs. 
The interpretations were generated by IG \citep{sundararajan2017axiomatic}, which visualizes the attribution of each input feature (word/token) to the model prediction. 
Figure \ref{fig:illustrations} (1) shows the predictions and interpretations of model A, B, and C on a \textsc{positive} movie review and its adversarial counterpart. 
Model A is not robust as its prediction on the adversarial example is flipped and the interpretation is totally changed. 
Although model B makes the same predictions on the original and adversarial examples, its interpretations reveal that these predictions are based on different key features: for the original example, it is a sentiment word \texttt{clever}; for the adversarial example, it is a neutral word \texttt{cinema}. 
The interpretation discrepancy reveals the vulnerability of model B, as shown in Fig. \ref{fig:illustrations} (2), where we craft another adversarial attack. 
Model B fails to recognize \texttt{dull} and \texttt{pesky} as the same important, and makes a wrong prediction on the \textsc{negative} adversarial example based on \texttt{cinema}.  
Only model C is robust as it behaves consistently on predicting both original/adversarial example pairs. 
Note that we look at model robustness through the lens of interpretations, while leaving the problem of trustworthiness or robustness of an interpretation method itself out as that is beyond the scope of this paper.

Based on the previous discussion, we argue that a robust model should have consistent prediction behaviors on original/adversarial example pairs, that is making the same predictions (\textbf{\textit{what}}) based on the same reasons (\textbf{\textit{how}}) which are reflected by consistent interpretations, as the word saliency maps of model C in Fig. \ref{fig:illustrations}. 
However, traditional adversarial training does not regularize model prediction behavior for improving model robustness. 
To train a robust model, we propose a fine-grained feature-level adversarial training named \ourmethod. 
\ourmethod learns global word importance via variational word masks \citep{chen2020learning} and regularizes the importance scores of the replaced words and their substitutions in original/adversarial example pairs during training. 
\ourmethod teaches the model to behave consistently on predicting original/adversarial example pairs by focusing on the corresponding important words based on their importance scores, hence improving the model robustness to adversarial examples.

The contribution of this work is three-fold: (1) we argue that adversarial training should improve model robustness by making the model produce the same predictions on original/adversarial example pairs with consistent interpretations; (2) we propose a new training strategy, feature-level adversarial training (\ourmethod), to achieve this goal by regularizing model prediction behaviors on original/adversarial example pairs to be consistent; and (3) we evaluate the effectiveness of \ourmethod in improving the robustness of four neural network models, LSTM \citep{hochreiter1997long}, CNN \citep{kim2014convolutional}, BERT \citep{devlin2018bert}, and DeBERTa \citep{he2020deberta}, to two adversarial attacks on four text classification tasks. The models trained via FLAT also show better robustness than baseline models on unforeseen adversarial examples across six different attacks.

\section{Related Work}
\label{sec:relate}
Neural language models have shown vulnerability to adversarial examples which are generated by manipulating input texts, such as replacing words with their synonyms \citep{alzantot2018generating, ren2019generating, jin2020bert, li2020contextualized, garg-ramakrishnan-2020-bae}, introducing character typos \citep{li2018textbugger, gao2018black}, paraphrasing sentences \citep{ribeiro-etal-2018-semantically, iyyer-etal-2018-adversarial}, and inserting malicious triggers \citep{wallace2019universal}. 
In this work, we focus on word substitution-based attacks as the generated adversarial examples largely maintain the original semantic meaning and lexical and grammatical correctness compared to other attacks \citep{zhang2020adversarial}. 
Some methods defend this kind of attacks by detecting malicious inputs and preventing them from attacking a model \citep{zhou-etal-2019-learning, mozes-etal-2021-frequency, wang2019sem}. 
However, blocking adversaries does not essentially solve the vulnerability problem of the target model. 
Another line of works focus on improving model robustness, and broadly fall into two categories: (1) certifiably robust training and (2) adversarial training.

\paragraph{Certifiably robust training.}
\citet{jia2019certified, huang-etal-2019-achieving} utilized interval bound propagation (IBP) to bound model robustness. \citet{shi2020robustness, xu2020automatic} extended the robustness verification to transformers, while it is challenging to scale these methods to complex neural networks (e.g. BERT) without loosening bounds. \citet{ye-etal-2020-safer} proposed a structure-free certified robust method which can be applied to advanced language models. However, the certified accuracy is at the cost of model performance on clean data.

\paragraph{Adversarial training.}
Adversarial training in text domain usually follows two steps: (1) collecting adversarial examples by attacking a target model and (2) fine-tuning the model on the augmented dataset with these adversarial examples. The augmented adversarial examples can be real examples generated by perturbing the original texts \citep{jin2020bert, li2020contextualized}, produced by generation models \citep{wang-etal-2020-cat}, or virtual examples crafted from word embedding space by adding noise to original word embeddings \citep{miyato2016adversarial, zhu2019freelb} or searching the worst-case in a convex hull \citep{zhou-etal-2021-defense, dong2021towards}. 

The above methods improve model robustness by solely looking at model predictions, that is making a model produce the same correct predictions on original/adversarial example pairs. Nevertheless, a robust model should behave consistently on predicting similar texts beyond producing the same predictions. Regularizing model prediction behavior should be considered in improving model robustness, no matter via certifiably robust training or adversarial training. In this work, we focus on extending traditional adversarial training to fine-grained feature-level adversarial training (\ourmethod), while leaving adding constraints on model behavior in certifiably robust training to future work. Besides, \ourmethod is compatible with existing substitution-based adversarial data augmentation methods. We focus on those \citep{jin2020bert, ren2019generating} that generate adversarial examples by perturbing original texts in our experiments. 

Another related work bounds model robustness by regularizing interpretation discrepancy between original and adversarial examples in image domain \citep{boopathy2020proper}. However, the interpretations are post-hoc and could vary across different interpretation methods. Differently, \ourmethod learns global feature importance during training. By regularizing the global importance of replaced words and their substitutions, the model trained via \ourmethod would be robust to unforeseen adversarial examples in which the substitution words appear.

\section{Method}
\label{sec:method}

\begin{figure}[t]
  \centering
  \includegraphics[width=0.46\textwidth]{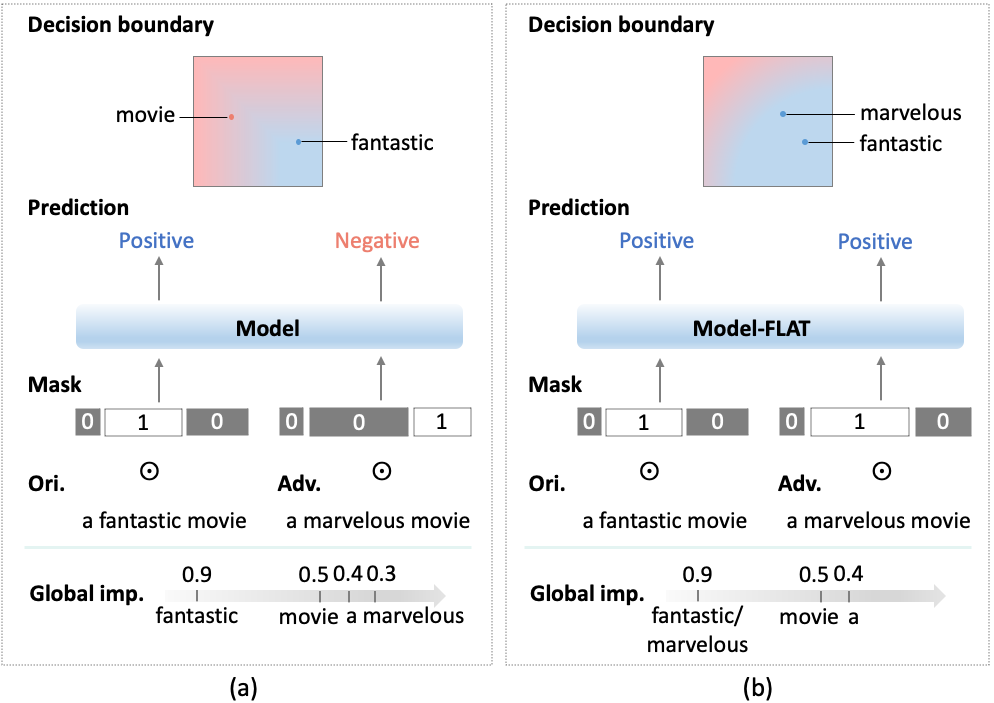}
  \caption{\label{fig:framework} (a) The model with variational word masks trained on the standard training set. As \texttt{marvelous} is not recognized as the same important as its synonym \texttt{fantastic} and masked out, the model makes a wrong prediction based on a neutral word \texttt{movie}. (b) \ourmethod increases the global importance of \texttt{marvelous} and teaches the model to make the same correct predictions on the original/adversarial example pair by focusing on \texttt{fantastic} and \texttt{marvelous} respectively.}
\end{figure}

This section introduces the proposed \ourmethod method. 
\ourmethod aims at improving model robustness by making a model behave consistently on predicting original/adversarial example pairs. 
To achieve this goal, \ourmethod leverages variational word masks to select the corresponding words (e.g. \texttt{fantastic} and \texttt{marvelous} in Fig. \ref{fig:framework}) from an original/adversarial example pair for the model to make predictions. 
To ensure the correctness of model predictions, variational word masks learn global word importance during training and play as a bottleneck teaching the model to make predictions based on important words. 
Besides, \ourmethod regularizes the global importance of the replaced words in an original example and their substitutions in the adversarial counterpart so that the model would recognize the corresponding words as the same important (or unimportant), as Fig. \ref{fig:framework} shows. 

\paragraph{Preliminaries.}
Given an input $\vec{x}=[\vec{x}_{1}, \ldots, \vec{x}_{n}]$, where $\vec{x}_{i}\in \mathbb{R}^{d}$ ($i \in \{1, \dots, n\}$) denotes the word embedding, the model $f_{\vec{\theta}}(\cdot)$ with parameter $\vec{\theta}$ outputs a prediction label $y=f_{\vec{\theta}}(\vec{x})$ for text classification tasks. An adversarial example $\vec{x}'$ is crafted from $\vec{x}$ under some constraints, such as maintaining the original semantic meaning. For word substitution-based adversarial attacks, an adversarial example replaces some words $\{\vec{x}_{i}\}$ in the original example $\vec{x}$ with their synonyms $\{\vec{x}'_{i}\}$. The adversarial example fools the model to output a different label, i.e. $y'=f_{\vec{\theta}}(\vec{x}')\neq y$.

We obtain a set of adversarial examples $\mathcal{D}'=\{(\vec{x}'^{(m)}, y^{(m)})\}$ by attacking the model on the original dataset $\mathcal{D}=\{(\vec{x}^{(m)}, y^{(m)})\}$. During adversarial training, the model is trained on both original and adversarial examples ($\mathcal{D} \cup \mathcal{D}'$). In addition to improving model prediction accuracy on adversarial examples, adversarial training should also make the model produce the same predictions on the similar texts with consistent decision-makings. Failing to do this would make the model vulnerable to unforeseen adversarial examples crafted with the substitution words in some other contexts. To achieve this goal, we propose the feature-level adversarial training (\ourmethod) method.

\subsection{Feature-Level Adversarial Training}
\label{sec:fea_adv_train}

Recall the goal of \ourmethod is to train a robust model with consistent prediction behaviors on original/adversarial example pairs. There are two desiderata for \ourmethod:
\begin{enumerate}
	\item Global feature importance scores $\vec{\phi}$. To teach the model to recognize the replaced words in an original example and their substitutions in the adversarial counterpart as the same important (or unimportant) for predictions, \ourmethod needs to learn the global importance score $\phi_{\vec{x}_{i}}$ of a word $\vec{x}_{i}$. Note that the ``global'' means the importance score is solely dependent on the word (embedding).
	\item Feature selection function $g_{\vec{\phi}}(\cdot)$. To guide the model to make predictions based on the corresponding important words in the original and adversarial example respectively, \ourmethod needs a feature selection function $g_{\vec{\phi}}(\cdot)$. $g_{\vec{\phi}}(\vec{x})$ selects important words from $\vec{x}$ based on their global importance scores in $\vec{\phi}$. The selected words are then forwarded to the model to output a prediction, i.e. $y=f_{\vec{\theta}}(g_{\vec{\phi}}(\vec{x}))$.
\end{enumerate}
\ourmethod leverages variational word masks \citep{chen2020learning} to learn global feature importance scores and select important features for model predictions, which will be introduced in Section \ref{sec:fea_select}.

With the two desiderata, the objective of \ourmethod is formulated as
\begin{eqnarray}
\min_{\vec{\theta}, \vec{\phi}} &\mathcal{L}_{pred} + \gamma \mathcal{L}_{imp} \label{eq:obj_1}\\
\mathcal{L}_{pred}&=\mathop{\mathbb{E}}_{(\vec{x}, y)\sim \mathcal{D}}[\mathcal{L}(f_{\vec{\theta}}(g_{\vec{\phi}}(\vec{x})), y)] \label{eq:pred_loss} \\
&+ \mathop{\mathbb{E}}_{(\vec{x}', y)\sim \mathcal{D}'}[\mathcal{L}(f_{\vec{\theta}}(g_{\vec{\phi}}(\vec{x}')), y)] \nonumber \\
\mathcal{L}_{imp}&=\mathop{\mathbb{E}}_{(\vec{x}, \vec{x}')\sim \mathcal{D} \cup \mathcal{D}'}[\sum_{i, \vec{x}_{i} \neq \vec{x}'_{i}} \lvert \phi_{\vec{x}_{i}} - \phi_{\vec{x}'_{i}} \rvert] \label{eq:imp_loss}
\end{eqnarray}
where $\mathcal{L}(\cdot, \cdot)$ denotes cross entropy loss. $\vec{\phi}$ is a learnable vector with the same dimension as the predefined vocabulary, where $\phi_{\vec{x}_{i}} \in (0, 1)$ represents the global importance of the word $\vec{x}_{i}$. $\gamma \in\mathbb{R}_{+}$ is a coefficient. $\mathcal{L}_{imp}$ regularizes the global importance scores of the replaced words $\{\vec{x}_{i}\}$ and their substitutes $\{\vec{x}'_{i}\}$ in an original/adversarial example pair $(\vec{x}, \vec{x}')$ by pushing $\phi_{\vec{x}_{i}}$ and $\phi_{\vec{x}'_{i}}$ close. With similar importance scores, the associated word pair $(\vec{x}_{i}, \vec{x}'_{i})$ would be selected by $g_{\vec{\phi}}(\cdot)$ or not simultaneously. $\mathcal{L}_{pred}$ encourages the model to make the same and correct predictions on the original and adversarial example based on the selected important words $g_{\vec{\phi}}(\vec{x})$ and $g_{\vec{\phi}}(\vec{x}')$ respectively. By optimizing the objective, the model learns to behave consistently on predicting similar texts, hence having better robustness to adversarial attacks.

\subsection{Learning with Variational Word Masks}
\label{sec:fea_select}

\ourmethod fulfills the two desiderata by training the model ($f_{\vec{\theta}}(\cdot)$) with variational word masks \citep{chen2020learning}. Specifically, variational word masks learn global word importance $\vec{\phi}$ during training and select important words for the model to make predictions by masking out irrelevant or unimportant words. For an input $\vec{x}=[\vec{x}_{1}, \ldots, \vec{x}_{n}]$, a set of masks $\vec{W}=[W_{\vec{x}_{1}}, \ldots, W_{\vec{x}_{n}}]$ are generated based on $\vec{\phi}$, where $W_{\vec{x}_{i}} \in \{0, 1\}$ is a binary random variable with 0 and 1 indicating to mask out or select the word $\vec{x}_{i}$ respectively. The word importance score $\phi_{\vec{x}_{i}}$ is the expectation of $W_{\vec{x}_{i}}$, that is the probability of the word $\vec{x}_{i}$ being selected. The feature selection function $g_{\vec{\phi}}(\cdot)$ in Section \ref{sec:fea_adv_train} is defined as 
\begin{equation}
\label{eq:fea_select}
g_{\vec{\phi}}(\vec{x})=\vec{W} \odot \vec{x},
\end{equation}
where $\odot$ denotes element-wise multiplication.

To ensure the model concentrating on a few important words to make predictions, we regularize $\vec{W}$ by maximizing its entropy conditioned on $\vec{x}$. The intuition is that most words in the vocabulary are irrelevant or noisy features (e.g. stop words) to text classification tasks \citep{chen2020learning}. The regularization on $\vec{W}$ will push the importance scores of most irrelevant words close to 0.5, while making a few important words have relatively high importance scores (close to 1), and the rest unimportant words have low scores (close to 0). Under this constraint, we rewrite the prediction loss $\mathcal{L}_{pred}$ in the objective (\ref{eq:obj_1}) as
\begin{eqnarray}
\label{eq:pred_loss1}
\mathcal{L}_{pred}=\mathop{\mathbb{E}}_{(\vec{x}, y)\sim \mathcal{D}}[\mathbb{E}_{q}[\mathcal{L}(f_{\vec{\theta}}(\vec{W} \odot \vec{x}), y)] -\beta H_{q}(\vec{W} \mid \vec{x})] \nonumber\\
+\mathop{\mathbb{E}}_{(\vec{x}', y)\sim \mathcal{D}'}[\mathbb{E}_{q'}[\mathcal{L}(f_{\vec{\theta}}(\vec{W}' \odot \vec{x}'), y)] -\beta H_{q'}(\vec{W}' \mid \vec{x}')],\nonumber
\end{eqnarray}
where $q=q_{\vec{\phi}}(\vec{W} \mid \vec{x})$ and $q'=q_{\vec{\phi}}(\vec{W}' \mid \vec{x}')$ denote the distributions of word masks on the original example $\vec{x}$ and adversarial example $\vec{x}'$ respectively, $H_{q}(\cdot\mid\cdot)$ is the conditional entropy, and $\beta \in\mathbb{R}_{+}$ is a coefficient.

\subsection{Connection}
\label{sec:connect}
\ourmethod degrades to traditional adversarial training when all words are regarded as equal important (all mask values are 1), and no constraint is added to regularize the importance scores of associated words in original/adversarial example pairs. Traditional adversarial training simply updates the model on the augmented dataset $\mathcal{D} \cup \mathcal{D}'$ by optimizing
\begin{equation}
\label{eq:tra_obj}
\min_{\vec{\theta}} \!\!\mathop{\mathbb{E}}_{(\vec{x}, y)\sim \mathcal{D}}\!\![\mathcal{L}(f_{\vec{\theta}}(\vec{x}), y)] + \!\!\mathop{\mathbb{E}}_{(\vec{x}', y)\sim \mathcal{D}'}\!\![\mathcal{L}(f_{\vec{\theta}}(\vec{x}')), y)].
\end{equation}
With no constraint on model prediction behavior on predicting similar texts, the model robustness is not guaranteed, especially to unforeseen adversarial attacks, as the results shown in experiments.

\subsection{Implementation Specification}
\label{sec:implement}

We utilize the amortized variational inference \citep{kingma2013auto} to approximate word mask distributions, and learn the parameter $\vec{\phi}$ (global word importance) via an inference network which is a single-layer feedforward neural network. For simplicity, we assume the word masks are mutually independent and each mask is dependent on the word embedding, that is $q_{\vec{\phi}}(\vec{W} \mid \vec{x})=\prod_{i=1}^{n}q_{\vec{\phi}}(W_{\vec{x}_{i}} \mid \vec{x}_{i})$. We optimize the inference network with the model jointly via stochastic gradient descent, and apply the Gumbel-softmax trick \citep{jang2016categorical, maddison2016concrete} to address the discreteness of sampling binary masks from Bernoulli distributions in backpropagation \citep{chen2020learning}. In the inference stage, we multiply each word embedding and its global importance score for the model to make predictions.

We first train a base model on the original dataset, and attack the model by manipulating the original training data and collect adversarial examples. Then we train the model on both original and adversarial examples via \ourmethod. We repeat the attacking and training processes 3-5 times (depending on the model and dataset) until convergence. Note that in each iteration, we augment the original training data with new adversarial examples generated by attacking the latest checkpoint.

\section{Experimental Setup}
\label{sec:setup}
The proposed method is evaluated with four neural network models in defending two adversarial attacks on four text classification tasks.

 \begin{table}
 	\centering
 	\small
 	\begin{tabular}{cccccc}
 		\toprule
 		Datasets & \textit{C} & \textit{L} & \textit{\#train} & \textit{\#dev} & \textit{\#test} \\
 		\midrule
 		SST2 & 2 & 19 & 6920 & 872 & 1821 \\
 		IMDB & 2& 268 & 20K & 5K & 25K  \\
 		AG & 4 & 32 & 114K & 6K  & 7.6K \\
 		TREC & 6 & 10 & 5000 & 452  & 500 \\
 		\bottomrule
 	\end{tabular}
 	\caption{Summary statistics of the datasets, where \textit{C} is the number of classes, \textit{L} is the average sentence length, \textit{\#} counts the number of examples in the \textit{train/dev/test} sets.}
 	\label{tab:datasets}
 \end{table}

\paragraph{Datasets.}
The four text classification datasets are:  Stanford Sentiment Treebank with binary labels SST2 \citep{socher2013recursive}, movie reviews IMDB \citep{maas2011learning}, AG’s News (AG) \citep{zhang2015character}, and 6-class question classification TREC \citep{li2002learning}. For the datasets (e.g. IMDB) without standard train/dev/test split, we hold out a proportion of training examples as the development set. Table \ref{tab:datasets} shows the statistics of the datasets.

\paragraph{Models.} 
We evaluate the proposed method with a recurrent neural network \citep[LSTM]{hochreiter1997long}, a convolutional neural network \citep[CNN]{kim2014convolutional}, and two state-of-the-art transformer-based models—BERT \citep{devlin2018bert} and DeBERTa \citep{he2020deberta}. The LSTM and CNN are initialized with 300-dimensional pretrained word embeddings~\cite{mikolov2013distributed}. We adopt the base versions of both BERT and DeBERTa.

\paragraph{Attack methods.}
We adopt two adversarial attacks, Textfooler \citep{jin2020bert} and PWWS \citep{ren2019generating}. Both methods check the lexical correctness and semantic similarity of adversarial examples with their original counterparts. The adversarial attacks are conducted on the TextAttack benchmark \citep{morris2020textattack} with default settings. During adversarial training, we attack all training data for the SST2 and TREC datasets to collect adversarial examples, while randomly attacking 10K training examples for the IMDB and AG datasets due to computational costs. 

More details of experimental setup are in Appendix \ref{sec:sup_exp}.

\begin{table}[t] 
	\small
	\centering
	\begin{tabular}{p{3.8cm}p{0.6cm}p{0.6cm}P{0.6cm}p{0.6cm}}
		\toprule
		Models & SST2 & IMDB & AG & TREC \\
		\midrule
		LSTM-base & 84.40  & 88.03  & 91.08  &  90.80 \\
		LSTM-adv(Textfooler) & 82.32  & 88.79  &  90.29 & 87.60 \\
		LSTM-adv(PWWS) &  82.59 & 88.37  & 91.16  & 89.60 \\
		LSTM-\ourmethod(Textfooler) & \textbf{84.79}  & \textbf{89.17}  & 91.00  & 91.00 \\
		LSTM-\ourmethod(PWWS) & 83.69  & 88.52  & \textbf{91.37}  & \textbf{91.20} \\
		\midrule
		CNN-base & \textbf{84.18}  & 88.63  & 91.32  &  \textbf{91.20} \\
		CNN-adv(Textfooler) & 82.15  & 88.81  & 90.99  & 89.20 \\
		CNN-adv(PWWS) & 83.42  & 88.89  & 91.30  & 90.00 \\
		CNN-\ourmethod(Textfooler) & 83.09  & 88.89  & \textbf{91.64}  & 89.20 \\
		CNN-\ourmethod(PWWS) & 83.31  & \textbf{88.99}  & 91.03  & 89.20 \\
		\midrule
		BERT-base & 91.32  & 91.71  & 93.59  & \textbf{97.40} \\
		BERT-adv(Textfooler) & 91.38  & 92.50  & 90.30  & 96.00 \\
		BERT-adv(PWWS) & 90.88  & \textbf{93.14}  & 93.38  & 95.20 \\
		BERT-\ourmethod(Textfooler) & \textbf{91.54}  &  92.78 & \textbf{94.07}  & 96.20 \\
		BERT-\ourmethod(PWWS) & 91.05  & 93.11  & 93.09  & 96.60\\
		\midrule
		DeBERTa-base & 94.18  & 93.80  & 93.62  & 96.40 \\
		DeBERTa-adv(Textfooler) & 94.40  & 92.86  & 92.84  & 95.60 \\
		DeBERTa-adv(PWWS) & \textbf{94.78}  & 94.17  & 92.96  & 96.40 \\
		DeBERTa-\ourmethod(Textfooler) & 94.29  & \textbf{94.29}  & \textbf{94.29}  & 96.40 \\
		DeBERTa-\ourmethod(PWWS) &  94.12 & 94.26  & 93.82 & 96.40 \\
		\bottomrule
	\end{tabular}
	\caption{Prediction accuracy (\%) of different models on standard test sets.}
	\label{tab:ori-acc}
\end{table}

\section{Results}
\label{sec:results}

We train the four models on the four datasets with different training strategies. The base model trained on the clean data is named with suffix ``-base''. The model trained via traditional adversarial training is named with suffix ``-adv''. The model trained via the proposed method is named with suffix ``-\ourmethod''. For fairness, traditional adversarial training repeats the attacking and training processes the same times as \ourmethod. Table \ref{tab:ori-acc} shows the prediction accuracy of different models on standard test sets. The attack method used for generating adversarial examples during training is noted in brackets. For example, ``CNN-\ourmethod(Textfooler)'' means the CNN model trained via \ourmethod with adversarial examples generated by Textfooler attack. Different from previous defence methods \citep{jones-etal-2020-robust, zhou-etal-2021-defense} that hurt model performance on clean data, adversarial training (``adv'' and ``\ourmethod'') does not cause significant model performance drop, and even improves prediction accuracy in some cases. Besides, we believe that producing high-quality adversarial examples for model training would further improve model prediction performance, and leave this to our future work. The rest of this section will focus on evaluating model robustness from both prediction and interpretation perspectives. The evaluation results are recorded in Table \ref{tab:robustness}.

\begin{table*}[h] 
	\small
	\centering
	\begin{tabular}{cccccccccccccc}
		\toprule
		 & & \multicolumn{3}{c}{SST2} & \multicolumn{3}{c}{IMDB} & \multicolumn{3}{c}{AG} & \multicolumn{3}{c}{TREC} \\
		\cmidrule(lr){3-5} \cmidrule(lr){6-8}\cmidrule(lr){9-11} \cmidrule(lr){12-14}
		 Attacks & Models & AA & KT & TI & AA & KT & TI & AA & KT & TI & AA & KT & TI \\
		\midrule
		\multirow{9}{*}{Textfooler} & LSTM-base & 5.05 & 0.46 & 0.68  & 0.16 & 0.53  & 0.46 & 45.00 & 0.76  & 0.81  & 44.40 & 0.62  & 0.89 \\
		& LSTM-adv & 12.36 & 0.49 & 0.68  & 29.18 & 0.60  & 0.58 & 48.39 & 0.76  &  0.82 & 51.20 & 0.51  & 0.87 \\
		& LSTM-\ourmethod & \textbf{17.76} & \textbf{0.58} & \textbf{0.75}  &  \textbf{31.38} & \textbf{0.66} & \textbf{0.65} & \textbf{54.16} & \textbf{0.82}  & \textbf{0.86}  &  \textbf{55.20} & \textbf{0.68} & \textbf{0.90} \\
		\cmidrule(lr){2-14}
		& CNN-base & 1.98 & 0.46  & 0.69  & 3.72 & 0.64  & 0.56 & 8.74 & 0.55  & 0.62  & 45.20 & 0.68  & 0.91 \\
		& CNN-adv & 2.53 & 0.52  & 0.72  & 16.04 & 0.71  & 0.65 & 15.84 & 0.55 & 0.62  & 52.60 & 0.71  & 0.92 \\
		& CNN-\ourmethod &  \textbf{37.07} & \textbf{0.70} & \textbf{0.82}  & \textbf{32.62} & \textbf{0.76}  & \textbf{0.75} & \textbf{25.18} & \textbf{0.61}  & \textbf{0.67}  &  \textbf{62.20} & \textbf{0.87} & \textbf{0.96} \\
		\cmidrule(lr){2-14}
		& BERT-base & 4.72 & 0.35  & 0.56  & 3.84 & 0.38  & 0.33 &  11.84 & 0.39 &  0.48 & 37.60 & 0.44  & 0.87 \\
		& BERT-adv & 5.60 & 0.33  & 0.56  & 23.28 & 0.34  & 0.25 & 30.67 & 0.25  & 0.40  & 40.00 & 0.52  & 0.89 \\
		& BERT-\ourmethod &  \textbf{12.41} & \textbf{0.44} & \textbf{0.64}  &  \textbf{28.35} & \textbf{0.46} & \textbf{0.38} &  \textbf{32.29} & \textbf{0.45} & \textbf{0.53}  & \textbf{55.00} & \textbf{0.58} & \textbf{0.90} \\
		\cmidrule(lr){2-14}
		& DeBERTa-base & 5.22 & 0.64  & 0.76  & 2.82 & 0.71  & 0.72 & 12.12 & 0.60  & 0.63  &  39.00 & 0.69 & 0.92 \\
		& DeBERTa-adv & 7.96 & 0.60  & 0.73  & 8.38 & 0.81 & 0.77 & 25.70 & 0.61 & 0.62  & 42.80 & 0.69  & 0.93 \\
		& DeBERTa-\ourmethod &  \textbf{11.59} & \textbf{0.70} & \textbf{0.79}  & \textbf{24.62} & \textbf{0.83} & \textbf{0.78} &  \textbf{31.62} & \textbf{0.62} & \textbf{0.65}  & \textbf{49.60} & \textbf{0.73} & \textbf{0.94} \\
		\midrule
		\multirow{9}{*}{PWWS} & LSTM-base & 11.64 & 0.51 & 0.71  & 0.29 & 0.55  & 0.48 & 54.53 & 0.82  & 0.86  & 54.40 & 0.66  & 0.90 \\
		& LSTM-adv & 18.73 & 0.57 & 0.74  & 23.68 & 0.63  & 0.61 &  61.17 & 0.84 & 0.88 & 64.20 & 0.61  & 0.88 \\
		& LSTM-\ourmethod & \textbf{19.66} & \textbf{0.60} &  \textbf{0.75} & \textbf{25.00} & \textbf{0.69}  & \textbf{0.67} &  \textbf{62.41} & \textbf{0.85} & \textbf{0.89} & \textbf{67.80} & \textbf{0.79} & \textbf{0.94} \\
		\cmidrule(lr){2-14}
		& CNN-base & 8.29 & 0.53  & 0.72  & 4.36 & 0.72  & 0.59 & 18.86 & 0.57  & 0.64  &  54.20 & 0.71 & 0.91 \\
		& CNN-adv & 12.63 & 0.57  & 0.73  & 20.64 & 0.72  & 0.68 & 33.21 & 0.56  & 0.63  & 63.00 & 0.76  & 0.92 \\
		& CNN-\ourmethod & \textbf{14.83} & \textbf{0.58}  & \textbf{0.74} & \textbf{20.70} & \textbf{0.73}  & \textbf{0.69} &  \textbf{71.37} & \textbf{0.91} & \textbf{0.93}  & \textbf{65.60} & \textbf{0.77}  & \textbf{0.93} \\
		\cmidrule(lr){2-14}
		& BERT-base & 11.70 & 0.37  & 0.57  & 7.08 & 0.36  & 0.32 & 32.34 & 0.28  & 0.40  & 51.60 & 0.52  & 0.87 \\
		& BERT-adv & 14.44 & 0.37 & 0.58  & 18.32 & 0.33  & 0.29 & 33.38 & 0.29  & 0.40  & 65.20 & 0.45  & 0.86 \\
		& BERT-\ourmethod & \textbf{14.61} & \textbf{0.44} &  \textbf{0.64} & \textbf{25.08} & \textbf{0.41}  & \textbf{0.36} & \textbf{49.16} & \textbf{0.30}  & \textbf{0.42}  & \textbf{68.20} & \textbf{0.64}  & \textbf{0.90} \\
		\cmidrule(lr){2-14}
		& DeBERTa-base & 14.17 & 0.72  & 0.81  & 7.04 & 0.82  & 0.80 & 31.30 & 0.65  & 0.71  & 52.80 & 0.73  & 0.94 \\
		& DeBERTa-adv & 15.16 & 0.65  & 0.76  & 18.66 & 0.81 & 0.78 & 53.02 & 0.65  &  0.70 & 63.60 & 0.64  & 0.91 \\
		& DeBERTa-\ourmethod &  \textbf{23.23} & \textbf{0.75} & \textbf{0.83}  &  \textbf{26.58} & \textbf{0.84} & \textbf{0.81} &  \textbf{55.14} & \textbf{0.67} & \textbf{0.72}  & \textbf{66.40} & \textbf{0.80} & \textbf{0.95} \\
		\bottomrule
	\end{tabular}
	\caption{Model robustness to adversarial attacks in terms of predictions and interpretations. AA: after-attack accuracy (\%); KT: Kendall’s Tau order rank correlation; TI: top-k intersection ($k=5$).}
	\label{tab:robustness}
\end{table*}

\subsection{Prediction Robustness}
\label{sec:pred_robust}

We evaluate the prediction robustness of well-trained models by attacking them with adversarial examples crafted from original test examples. The model prediction accuracy on adversarial examples is denoted as after-attack accuracy \citep{jin2020bert}. In Table \ref{tab:robustness}, we omit the attack name in naming a model (``-adv'' or ``-\ourmethod'') as it is trained with adversarial examples generated by the corresponding attack method (Textfooler or PWWS).

Table \ref{tab:robustness} shows that base models are easily fooled by adversarial examples, achieving much lower after-attack accuracy than other models (``-\ourmethod'' and ``-adv'') trained with adversarial examples. \ourmethod consistently outperforms traditional adversarial training, indicating the effectiveness of regularizing model prediction behavior during adversarial training in improving prediction robustness. All the models show better prediction robustness on multiclass topic classification tasks (AG and TREC) than on binary sentiment classification tasks (SST2 and IMDB). Besides, the after-attack accuracy on the IMDB dataset is the lowest for most of the base models (especially LSTM-base). We suspect that IMDB has longer average text length than other datasets, which is easier to find successful adversarial examples. \ourmethod improves the after-attack accuracy of base models $15\%-30\%$ on the IMDB dataset.

\subsection{Interpretation Consistency}
\label{sec:interp_robust}

Beyond prediction robustness, model robustness can also be evaluated by comparing its decision-makings on predicting original/adversarial example pairs, i.e. interpretation consistency. Note that we obtain interpretations via local post-hoc interpretation methods that identify feature (word/token) attributions to the model prediction per example. We adopt two interpretation methods, IG \citep{sundararajan2017axiomatic} and LIME \citep{ribeiro2016should}, which are the representatives from two typical categories, white-box interpretations and black-box interpretations, respectively. IG computes feature attributions by integrating gradients of points along a path from a baseline to the input. LIME explains neural network predictions by fitting a local linear model with input perturbations and producing word attributions. For IG, we evaluate all test examples and their adversarial counterparts. For LIME, we randomly pick up 1000 example pairs for evaluation due to computational costs. We evaluate interpretation consistency under two metrics, Kendall’s Tau order rank correlation \citep{chen2019robust, boopathy2020proper} and Top-k intersection \citep{chen2019robust, ghorbani2019interpretation}.
For both metrics, we compute the interpretation consistency on corresponding labels and take the average over all classes as the overall consistency. Table \ref{tab:robustness} reports the results of IG interpretations. The results of LIME interpretations (Table \ref{tab:lime_inter} in Appendix \ref{sec:sup_quan_eva}) show similar tendency.

\paragraph{Kendall’s Tau order rank correlation.}
We adopt this metric to compare the overall rankings of word attributions between different interpretations. Higher Kendall’s Tau order rank correlation indicates better interpretation consistency. The models (``-\ourmethod'') outperform other baseline models (``-adv'' and ``-base'') with higher Kendall’s Tau order rank correlations, showing that \ourmethod teaches models to behave consistently on predicting similar texts. However, traditional adversarial training cannot guarantee the model robustness being improved as the interpretation discrepancy is even worse than that of base models in some cases, such as LSTM-adv and LSTM-base on the TREC dataset under the Textfooler attack. As \ourmethod consistently improves model interpretation consistency, no matter which interpretation method (IG or LIME) is used for evaluation, we believe the model robustness has been improved.

\paragraph{Top-k intersection.}
We adopt this metric to compute the proportion of intersection of top k important features identified by the interpretations of original/adversarial example pairs. Note that we treat synonyms as the "same" words. Higher top-k intersection indicates better interpretation consistency. Table \ref{tab:robustness} records the results of IG interpretations when $k=5$. The full results of top-k intersection with k increasing from 1 to 10 are in Appendix \ref{sec:sup_quan_eva}. Similar to the results of Kendall’s Tau order rank correlation, the models (``-\ourmethod'') outperform other baseline models (``-adv'' and ``-base'') with higher top-k intersection rates, showing that they tend to focus on the same words (or their synonyms) in original/adversarial example pairs to make predictions.

\section{Discussion}
\label{sec:discuss}

\begin{figure}[t]
	\centering
	\includegraphics[width=0.465\textwidth]{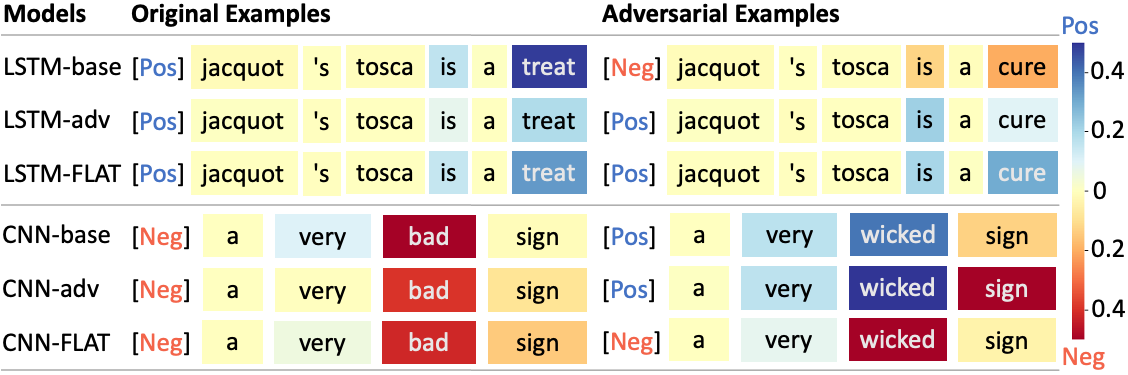}
	\caption{\label{fig:visualization} Visualization of IG interpretations. The model predictions are in ``[ ]''. The color of each block represents the word attribution to the model prediction.}
\end{figure}

\paragraph{Visualization of interpretations.}
Interpretations show the robustness of models (``-\ourmethod'') in producing the same predictions on original/adversarial example pairs with consistent decision-makings. 
Figure \ref{fig:visualization} visualizes the IG interpretations of LSTM- and CNN-based models on a \textsc{positive} and \textsc{negative} SST2 movie review respectively. The adversarial examples of the two movie reviews were generated by Textfooler. 
The base models (``-base'') were fooled by adversarial examples. Although LSTM-adv correctly predicted the \textsc{positive} original/adversarial example pair, its interpretations are discrepant with \texttt{treat} and \texttt{is} identified as the top important word respectively. For the \textsc{negative} adversarial example, CNN-adv failed to recognize \texttt{bad} and \texttt{wicked} as synonyms and labeled them with opposite sentiment polarities, which explains its wrong prediction. Both LSTM-\ourmethod and CNN-\ourmethod correctly predicted the original/adversarial example pairs with consistent interpretations.

\begin{table}[t] 
	\small
	\centering
	\begin{tabular}{p{2.2cm}p{0.6cm}p{0.5cm}P{0.5cm}p{0.5cm}p{0.5cm}p{0.5cm}}
		\toprule
		Models & PWWS & Gene & IGA & PSO & Clare & BAE \\
		\midrule
		LSTM-base &  11.64 & 20.26  & 9.83  &  5.88 & 3.02 & 36.52 \\
		LSTM-adv & 15.38  & 25.65  & 17.02  & 5.60 & 3.90 & 36.35 \\
		LSTM-\ourmethod & \textbf{20.48}  & \textbf{33.44}  &  \textbf{24.22} & \textbf{6.53} & \textbf{5.55} & \textbf{39.87} \\
		\midrule
		CNN-base & 8.29  & 20.32  & 7.85  & 5.60 & 1.48 & 37.12 \\
		CNN-adv & 8.68  & 16.42  & 6.26  & 5.60 & 1.04 & 35.48  \\
		CNN-\ourmethod &  \textbf{42.56} & \textbf{55.02}  & \textbf{46.35}  & \textbf{10.38} & \textbf{17.57} & \textbf{48.38} \\
		\midrule
		BERT-base & 11.70  & 32.24  & 9.72  & 6.26 & 0.86 & 35.31 \\
		BERT-adv & 13.01  & 34.49  & 10.87  & 6.64 & 1.04 & 36.74 \\
		BERT-\ourmethod & \textbf{15.93}  & \textbf{35.31}  & \textbf{15.93}  & \textbf{9.50} & \textbf{5.29} & \textbf{37.56} \\
		\midrule
		DeBERTa-base & 14.17  & 37.12  & 12.19  & 6.75 & 0.55 & 38.61 \\
		DeBERTa-adv & 17.52  & 37.18  & 12.85  & 7.96 & 1.07 & 40.14 \\
		DeBERTa-\ourmethod & \textbf{21.80}  & \textbf{48.16}  & \textbf{28.17}  & \textbf{13.01} & \textbf{1.37} & \textbf{44.54} \\
		\bottomrule
	\end{tabular}
	\caption{After-attack accuracy (\%) of different models to different attacks on the SST2 test set.}
	\label{tab:trans-acc}
\end{table}

\paragraph{Transferability of model robustness.}

The models trained via \ourmethod show better robustness than baseline models across different attacks. 
We test the robustness transferability of different models, where ``-adv'' and ``\ourmethod'' were trained with adversarial examples generated by Textfooler, to six unforeseen adversarial attacks: PWWS \citep{ren2019generating}, Gene \citep{alzantot2018generating}, IGA \citep{wang2019natural}, PSO \citep{zang-etal-2020-word}, Clare \citep{li2020contextualized}, and BAE \citep{garg-ramakrishnan-2020-bae}, which generate adversarial examples in different ways (e.g. WordNet swap \citep{miller1998wordnet}, BERT masked token prediction). The details of these attack methods are in Appendix \ref{sec:sup_exp}. Table \ref{tab:trans-acc} shows the after-attack accuracy of different models on the SST2 test set. The models trained via \ourmethod achieve higher after-attack accuracy than baseline models, showing better robustness to unforeseen adversarial examples. 

\paragraph{Ablation study.}
The regularizations on word masks and global word importance scores in the objective (\ref{eq:obj_1}) are important for improving model performance. 
We take the LSTM-\ourmethod model trained with Textfooler adversarial examples on the SST2 dataset for evaluation. The optimal hyperparameters are $\beta=0.1$, $\gamma=0.001$. We study the effects by setting $\beta$, $\gamma$, or both as zero. Table \ref{tab:ablation} shows the results. Only with both regularizations, the model can achieve good prediction performance on the clean test data (standard accuracy) and adversarial examples (after-attack accuracy). We observed that when $\beta=0$, all masks are close to 1, failing to learn feature importance. When $\gamma=0$, the model cannot recognize some words and their substitutions as the same important, which is reflected by the larger variance of L1 norm on the difference between the global importance of 1000 randomly sampled words and 10 of their synonyms, as Fig. \ref{fig:box_plot} shows in Appendix \ref{sec:sup_discussion}.

\begin{figure}
	\centering
	\includegraphics[width=0.47\textwidth]{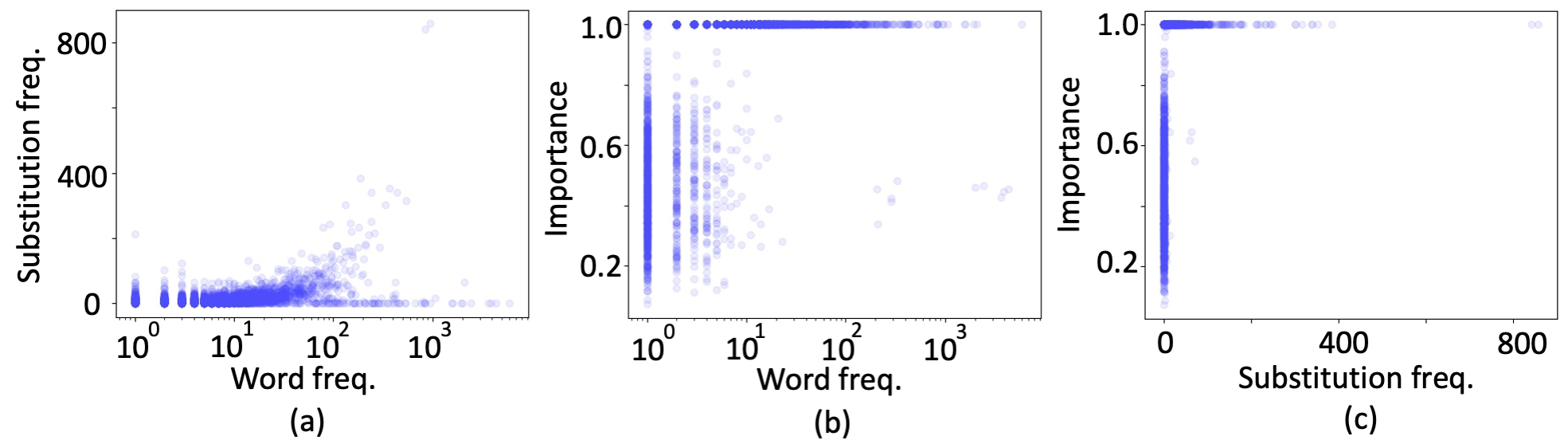}
	\caption{\label{fig:corr} Scatter plots: (a) substitution frequency vs. word frequency; (b) global importance vs. word frequency; (c) global importance vs. substitution frequency.}
\end{figure}

\begin{table}[h] 
	\small
	\centering
	\begin{tabular}{p{2.8cm}p{1.5cm}p{1.5cm}}
		\toprule
		Hyperparameters & SA & AA \\
		\midrule
		$\beta=0.1$, $\gamma=0.001$ & 84.79 & 17.76 \\
		$\beta=0.1$, $\gamma=0$ & 83.96 & 9.99 \\
		$\beta=0$, $\gamma=0.001$ & 84.34  & 8.18 \\
		$\beta=0$, $\gamma=0$ & 84.40  & 8.35 \\
		\bottomrule
	\end{tabular}
	\caption{The effects of \ourmethod regularizations on model performance. SA: standard accuracy (\%); AA: after-attack accuracy (\%)}
	\label{tab:ablation}
\end{table}

\paragraph{Correlations.}
The learned global word importance, word frequency, and word substitution frequency in adversarial examples do not show strong correlations with each other. 
We take the LSTM-\ourmethod trained with Textfooler on the SST2 dataset for analysis. 
As the scatter plots in Fig. \ref{fig:corr} show, any two of the three do not have strong correlations (as their Pearson correlations show in Table \ref{tab:correlation} in Appendix \ref{sec:sup_discussion}). 
Figure \ref{fig:corr} (a) shows that the replaced words are not based on their frequency. 
Figure \ref{fig:corr} (b) and (c) show that global word importance scores were learned during training, not trivially based on word frequency or substitution frequency. 
It is expected the words that have high substitution frequency in adversarial examples have high importance scores. In addition, \ourmethod also identifies some important words that are low-frequency or even not replaced by adversarial examples.

\section{Conclusion}
\label{sec:conclusion}
In this paper, we look into the robustness of neural network models from both prediction and interpretation perspectives. We propose a new training strategy, \ourmethod, to regularize a model prediction behavior so that it produces the same predictions on original/adversarial example pairs with consistent interpretations. We test \ourmethod with four neural network models, LSTM, CNN, BERT, and DeBERTa, and show its effectiveness in improving model robustness to two adversarial attacks on four text classification tasks.

\clearpage
\section*{Acknowledgments}

We thank the authors of \citet{morris2020textattack} for providing the TextAttack benchmark. We thank the anonymous reviewers for many valuable comments.
\bibliography{ref}

\begin{thebibliography}{53}
\providecommand{\natexlab}[1]{#1}

\bibitem[{Alzantot et~al.(2018)Alzantot, Sharma, Elgohary, Ho, Srivastava, and
  Chang}]{alzantot2018generating}
Alzantot, M.; Sharma, Y.; Elgohary, A.; Ho, B.-J.; Srivastava, M.; and Chang,
  K.-W. 2018.
\newblock Generating Natural Language Adversarial Examples.
\newblock In \emph{EMNLP}.

\bibitem[{Boopathy et~al.(2020)Boopathy, Liu, Zhang, Liu, Chen, Chang, and
  Daniel}]{boopathy2020proper}
Boopathy, A.; Liu, S.; Zhang, G.; Liu, C.; Chen, P.-Y.; Chang, S.; and Daniel,
  L. 2020.
\newblock Proper Network Interpretability Helps Adversarial Robustness in
  Classification.
\newblock In \emph{ICML}, 1014--1023. PMLR.

\bibitem[{Chen and Ji(2020)}]{chen2020learning}
Chen, H.; and Ji, Y. 2020.
\newblock Learning Variational Word Masks to Improve the Interpretability of
  Neural Text Classifiers.
\newblock In \emph{EMNLP}.

\bibitem[{Chen et~al.(2019)Chen, Wu, Rastogi, Liang, and Jha}]{chen2019robust}
Chen, J.; Wu, X.; Rastogi, V.; Liang, Y.; and Jha, S. 2019.
\newblock Robust attribution regularization.
\newblock In \emph{NeurIPS}.

\bibitem[{Devlin et~al.(2019)Devlin, Chang, Lee, and
  Toutanova}]{devlin2018bert}
Devlin, J.; Chang, M.-W.; Lee, K.; and Toutanova, K. 2019.
\newblock Bert: Pre-training of deep bidirectional transformers for language
  understanding.
\newblock In \emph{NAACL-HLT}.

\bibitem[{Dong et~al.(2021)Dong, Luu, Ji, and Liu}]{dong2021towards}
Dong, X.; Luu, A.~T.; Ji, R.; and Liu, H. 2021.
\newblock Towards robustness against natural language word substitutions.
\newblock In \emph{ICLR}.

\bibitem[{Dong, Dong, and Hao(2010)}]{dong-etal-2010-hownet}
Dong, Z.; Dong, Q.; and Hao, C. 2010.
\newblock {H}ow{N}et and Its Computation of Meaning.
\newblock In \emph{Coling 2010: Demonstrations}.

\bibitem[{Gao et~al.(2018)Gao, Lanchantin, Soffa, and Qi}]{gao2018black}
Gao, J.; Lanchantin, J.; Soffa, M.~L.; and Qi, Y. 2018.
\newblock Black-box generation of adversarial text sequences to evade deep
  learning classifiers.
\newblock In \emph{2018 IEEE SPW}.

\bibitem[{Garg and Ramakrishnan(2020)}]{garg-ramakrishnan-2020-bae}
Garg, S.; and Ramakrishnan, G. 2020.
\newblock {BAE}: {BERT}-based Adversarial Examples for Text Classification.
\newblock In \emph{EMNLP}.

\bibitem[{Ghorbani, Abid, and Zou(2019)}]{ghorbani2019interpretation}
Ghorbani, A.; Abid, A.; and Zou, J. 2019.
\newblock Interpretation of neural networks is fragile.
\newblock In \emph{AAAI}.

\bibitem[{He et~al.(2021)He, Liu, Gao, and Chen}]{he2020deberta}
He, P.; Liu, X.; Gao, J.; and Chen, W. 2021.
\newblock Deberta: Decoding-enhanced bert with disentangled attention.
\newblock In \emph{ICLR}.

\bibitem[{Hochreiter and Schmidhuber(1997)}]{hochreiter1997long}
Hochreiter, S.; and Schmidhuber, J. 1997.
\newblock Long short-term memory.
\newblock \emph{Neural computation}, 9(8): 1735--1780.

\bibitem[{Huang et~al.(2019)Huang, Stanforth, Welbl, Dyer, Yogatama, Gowal,
  Dvijotham, and Kohli}]{huang-etal-2019-achieving}
Huang, P.-S.; Stanforth, R.; Welbl, J.; Dyer, C.; Yogatama, D.; Gowal, S.;
  Dvijotham, K.; and Kohli, P. 2019.
\newblock Achieving Verified Robustness to Symbol Substitutions via Interval
  Bound Propagation.
\newblock In \emph{EMNLP-IJCNLP}.

\bibitem[{Iyyer et~al.(2018)Iyyer, Wieting, Gimpel, and
  Zettlemoyer}]{iyyer-etal-2018-adversarial}
Iyyer, M.; Wieting, J.; Gimpel, K.; and Zettlemoyer, L. 2018.
\newblock Adversarial Example Generation with Syntactically Controlled
  Paraphrase Networks.
\newblock In \emph{NAACL-HLT}.

\bibitem[{Jang, Gu, and Poole(2017)}]{jang2016categorical}
Jang, E.; Gu, S.; and Poole, B. 2017.
\newblock Categorical reparameterization with gumbel-softmax.
\newblock In \emph{ICLR}.

\bibitem[{Jia et~al.(2019)Jia, Raghunathan, G{\"o}ksel, and
  Liang}]{jia2019certified}
Jia, R.; Raghunathan, A.; G{\"o}ksel, K.; and Liang, P. 2019.
\newblock Certified Robustness to Adversarial Word Substitutions.
\newblock In \emph{EMNLP-IJCNLP}.

\bibitem[{Jin et~al.(2020)Jin, Jin, Zhou, and Szolovits}]{jin2020bert}
Jin, D.; Jin, Z.; Zhou, J.~T.; and Szolovits, P. 2020.
\newblock Is bert really robust? a strong baseline for natural language attack
  on text classification and entailment.
\newblock In \emph{AAAI}.

\bibitem[{Jones et~al.(2020)Jones, Jia, Raghunathan, and
  Liang}]{jones-etal-2020-robust}
Jones, E.; Jia, R.; Raghunathan, A.; and Liang, P. 2020.
\newblock Robust Encodings: A Framework for Combating Adversarial Typos.
\newblock In \emph{ACL}.

\bibitem[{Kim(2014)}]{kim2014convolutional}
Kim, Y. 2014.
\newblock Convolutional Neural Networks for Sentence Classification.
\newblock In \emph{EMNLP}.

\bibitem[{Kingma and Welling(2013)}]{kingma2013auto}
Kingma, D.~P.; and Welling, M. 2013.
\newblock Auto-encoding variational bayes.
\newblock \emph{arXiv preprint arXiv:1312.6114}.

\bibitem[{Li et~al.(2021)Li, Zhang, Peng, Chen, Brockett, Sun, and
  Dolan}]{li2020contextualized}
Li, D.; Zhang, Y.; Peng, H.; Chen, L.; Brockett, C.; Sun, M.-T.; and Dolan, B.
  2021.
\newblock Contextualized Perturbation for Textual Adversarial Attack.
\newblock In \emph{NAACL-HLT}.

\bibitem[{Li et~al.(2018)Li, Ji, Du, Li, and Wang}]{li2018textbugger}
Li, J.; Ji, S.; Du, T.; Li, B.; and Wang, T. 2018.
\newblock Textbugger: Generating adversarial text against real-world
  applications.
\newblock \emph{arXiv preprint arXiv:1812.05271}.

\bibitem[{Li and Roth(2002)}]{li2002learning}
Li, X.; and Roth, D. 2002.
\newblock Learning question classifiers.
\newblock In \emph{Proceedings of the 19th international conference on
  Computational linguistics-Volume 1}, 1--7. Association for Computational
  Linguistics.

\bibitem[{Liang et~al.(2017)Liang, Li, Su, Bian, Li, and Shi}]{liang2017deep}
Liang, B.; Li, H.; Su, M.; Bian, P.; Li, X.; and Shi, W. 2017.
\newblock Deep text classification can be fooled.
\newblock \emph{arXiv preprint arXiv:1704.08006}.

\bibitem[{Liu et~al.(2019)Liu, Ott, Goyal, Du, Joshi, Chen, Levy, Lewis,
  Zettlemoyer, and Stoyanov}]{liu2019roberta}
Liu, Y.; Ott, M.; Goyal, N.; Du, J.; Joshi, M.; Chen, D.; Levy, O.; Lewis, M.;
  Zettlemoyer, L.; and Stoyanov, V. 2019.
\newblock Roberta: A robustly optimized bert pretraining approach.
\newblock \emph{arXiv preprint arXiv:1907.11692}.

\bibitem[{Maas et~al.(2011)Maas, Daly, Pham, Huang, Ng, and
  Potts}]{maas2011learning}
Maas, A.~L.; Daly, R.~E.; Pham, P.~T.; Huang, D.; Ng, A.~Y.; and Potts, C.
  2011.
\newblock Learning word vectors for sentiment analysis.
\newblock In \emph{ACL}.

\bibitem[{Maddison, Mnih, and Teh(2016)}]{maddison2016concrete}
Maddison, C.~J.; Mnih, A.; and Teh, Y.~W. 2016.
\newblock The concrete distribution: A continuous relaxation of discrete random
  variables.
\newblock \emph{arXiv preprint arXiv:1611.00712}.

\bibitem[{Mikolov et~al.(2013)Mikolov, Sutskever, Chen, Corrado, and
  Dean}]{mikolov2013distributed}
Mikolov, T.; Sutskever, I.; Chen, K.; Corrado, G.~S.; and Dean, J. 2013.
\newblock Distributed representations of words and phrases and their
  compositionality.
\newblock In \emph{Advances in neural information processing systems},
  3111--3119.

\bibitem[{Miller(1998)}]{miller1998wordnet}
Miller, G.~A. 1998.
\newblock \emph{WordNet: An electronic lexical database}.
\newblock MIT press.

\bibitem[{Miyato, Dai, and Goodfellow(2017)}]{miyato2016adversarial}
Miyato, T.; Dai, A.~M.; and Goodfellow, I. 2017.
\newblock Adversarial training methods for semi-supervised text classification.
\newblock \emph{In Proceedings of the International Conference on Learning
  Representations}.

\bibitem[{Morris et~al.(2020)Morris, Lifland, Yoo, Grigsby, Jin, and
  Qi}]{morris2020textattack}
Morris, J.; Lifland, E.; Yoo, J.~Y.; Grigsby, J.; Jin, D.; and Qi, Y. 2020.
\newblock TextAttack: A Framework for Adversarial Attacks, Data Augmentation,
  and Adversarial Training in NLP.
\newblock In \emph{Proceedings of the 2020 Conference on Empirical Methods in
  Natural Language Processing: System Demonstrations}, 119--126.

\bibitem[{Mozes et~al.(2021)Mozes, Stenetorp, Kleinberg, and
  Griffin}]{mozes-etal-2021-frequency}
Mozes, M.; Stenetorp, P.; Kleinberg, B.; and Griffin, L. 2021.
\newblock Frequency-Guided Word Substitutions for Detecting Textual Adversarial
  Examples.
\newblock In \emph{Proceedings of the 16th Conference of the European Chapter
  of the Association for Computational Linguistics: Main Volume}, 171--186.
  Online: Association for Computational Linguistics.

\bibitem[{Mrk{\v{s}}i{\'c} et~al.(2016)Mrk{\v{s}}i{\'c}, S{\'e}aghdha, Thomson,
  Ga{\v{s}}i{\'c}, Rojas-Barahona, Su, Vandyke, Wen, and
  Young}]{mrkvsic2016counter}
Mrk{\v{s}}i{\'c}, N.; S{\'e}aghdha, D.~O.; Thomson, B.; Ga{\v{s}}i{\'c}, M.;
  Rojas-Barahona, L.; Su, P.-H.; Vandyke, D.; Wen, T.-H.; and Young, S. 2016.
\newblock Counter-fitting word vectors to linguistic constraints.
\newblock \emph{arXiv preprint arXiv:1603.00892}.

\bibitem[{Pennington, Socher, and Manning(2014)}]{pennington-etal-2014-glove}
Pennington, J.; Socher, R.; and Manning, C. 2014.
\newblock {G}lo{V}e: Global Vectors for Word Representation.
\newblock In \emph{EMNLP}.

\bibitem[{Ren et~al.(2019)Ren, Deng, He, and Che}]{ren2019generating}
Ren, S.; Deng, Y.; He, K.; and Che, W. 2019.
\newblock Generating natural language adversarial examples through probability
  weighted word saliency.
\newblock In \emph{ACL}.

\bibitem[{Ribeiro, Singh, and Guestrin(2016)}]{ribeiro2016should}
Ribeiro, M.~T.; Singh, S.; and Guestrin, C. 2016.
\newblock " Why should i trust you?" Explaining the predictions of any
  classifier.
\newblock In \emph{Proceedings of the 22nd ACM SIGKDD international conference
  on knowledge discovery and data mining}, 1135--1144.

\bibitem[{Ribeiro, Singh, and Guestrin(2018)}]{ribeiro-etal-2018-semantically}
Ribeiro, M.~T.; Singh, S.; and Guestrin, C. 2018.
\newblock Semantically Equivalent Adversarial Rules for Debugging {NLP} models.
\newblock In \emph{ACL}.

\bibitem[{Samanta and Mehta(2017)}]{samanta2017towards}
Samanta, S.; and Mehta, S. 2017.
\newblock Towards crafting text adversarial samples.
\newblock \emph{arXiv preprint arXiv:1707.02812}.

\bibitem[{Shi et~al.(2020)Shi, Zhang, Chang, Huang, and
  Hsieh}]{shi2020robustness}
Shi, Z.; Zhang, H.; Chang, K.-W.; Huang, M.; and Hsieh, C.-J. 2020.
\newblock Robustness verification for transformers.
\newblock \emph{arXiv preprint arXiv:2002.06622}.

\bibitem[{Socher et~al.(2013)Socher, Perelygin, Wu, Chuang, Manning, Ng, and
  Potts}]{socher2013recursive}
Socher, R.; Perelygin, A.; Wu, J.; Chuang, J.; Manning, C.~D.; Ng, A.; and
  Potts, C. 2013.
\newblock Recursive deep models for semantic compositionality over a sentiment
  treebank.
\newblock In \emph{EMNLP}.

\bibitem[{Sundararajan, Taly, and Yan(2017)}]{sundararajan2017axiomatic}
Sundararajan, M.; Taly, A.; and Yan, Q. 2017.
\newblock Axiomatic attribution for deep networks.
\newblock In \emph{International Conference on Machine Learning}, 3319--3328.
  PMLR.

\bibitem[{Wallace et~al.(2019)Wallace, Feng, Kandpal, Gardner, and
  Singh}]{wallace2019universal}
Wallace, E.; Feng, S.; Kandpal, N.; Gardner, M.; and Singh, S. 2019.
\newblock Universal adversarial triggers for attacking and analyzing NLP.
\newblock \emph{arXiv preprint arXiv:1908.07125}.

\bibitem[{Wang et~al.(2020)Wang, Wang, Qin, Packer, Li, Chen, Beutel, and
  Chi}]{wang-etal-2020-cat}
Wang, T.; Wang, X.; Qin, Y.; Packer, B.; Li, K.; Chen, J.; Beutel, A.; and Chi,
  E. 2020.
\newblock {CAT}-Gen: Improving Robustness in {NLP} Models via Controlled
  Adversarial Text Generation.
\newblock In \emph{EMNLP}.

\bibitem[{Wang, Jin, and He(2019)}]{wang2019natural}
Wang, X.; Jin, H.; and He, K. 2019.
\newblock Natural language adversarial attacks and defenses in word level.
\newblock \emph{arXiv preprint arXiv:1909.06723}.

\bibitem[{Wang et~al.(2021)Wang, Jin, Yang, and He}]{wang2019sem}
Wang, X.; Jin, H.; Yang, Y.; and He, K. 2021.
\newblock Natural Language Adversarial Defense through Synonym Encoding.
\newblock In \emph{UAI}.

\bibitem[{Xu et~al.(2020)Xu, Shi, Zhang, Wang, Chang, Huang, Kailkhura, Lin,
  and Hsieh}]{xu2020automatic}
Xu, K.; Shi, Z.; Zhang, H.; Wang, Y.; Chang, K.-W.; Huang, M.; Kailkhura, B.;
  Lin, X.; and Hsieh, C.-J. 2020.
\newblock Automatic perturbation analysis for scalable certified robustness and
  beyond.
\newblock \emph{Advances in Neural Information Processing Systems}, 33.

\bibitem[{Ye, Gong, and Liu(2020)}]{ye-etal-2020-safer}
Ye, M.; Gong, C.; and Liu, Q. 2020.
\newblock {SAFER}: A Structure-free Approach for Certified Robustness to
  Adversarial Word Substitutions.
\newblock In \emph{ACL}.

\bibitem[{Zang et~al.(2020)Zang, Qi, Yang, Liu, Zhang, Liu, and
  Sun}]{zang-etal-2020-word}
Zang, Y.; Qi, F.; Yang, C.; Liu, Z.; Zhang, M.; Liu, Q.; and Sun, M. 2020.
\newblock Word-level Textual Adversarial Attacking as Combinatorial
  Optimization.
\newblock In \emph{ACL}.

\bibitem[{Zhang et~al.(2020)Zhang, Sheng, Alhazmi, and
  Li}]{zhang2020adversarial}
Zhang, W.~E.; Sheng, Q.~Z.; Alhazmi, A.; and Li, C. 2020.
\newblock Adversarial attacks on deep-learning models in natural language
  processing: A survey.
\newblock \emph{ACM Transactions on Intelligent Systems and Technology (TIST)},
  11(3): 1--41.

\bibitem[{Zhang, Zhao, and LeCun(2015)}]{zhang2015character}
Zhang, X.; Zhao, J.; and LeCun, Y. 2015.
\newblock Character-level convolutional networks for text classification.
\newblock In \emph{Advances in neural information processing systems},
  649--657.

\bibitem[{Zhou et~al.(2019)Zhou, Jiang, Chang, and
  Wang}]{zhou-etal-2019-learning}
Zhou, Y.; Jiang, J.-Y.; Chang, K.-W.; and Wang, W. 2019.
\newblock Learning to Discriminate Perturbations for Blocking Adversarial
  Attacks in Text Classification.
\newblock In \emph{EMNLP-IJCNLP}.

\bibitem[{Zhou et~al.(2021)Zhou, Zheng, Hsieh, Chang, and
  Huang}]{zhou-etal-2021-defense}
Zhou, Y.; Zheng, X.; Hsieh, C.-J.; Chang, K.-W.; and Huang, X. 2021.
\newblock Defense against Synonym Substitution-based Adversarial Attacks via
  {D}irichlet Neighborhood Ensemble.
\newblock In \emph{ACL}.

\bibitem[{Zhu et~al.(2020)Zhu, Cheng, Gan, Sun, Goldstein, and
  Liu}]{zhu2019freelb}
Zhu, C.; Cheng, Y.; Gan, Z.; Sun, S.; Goldstein, T.; and Liu, J. 2020.
\newblock Freelb: Enhanced adversarial training for language understanding.
\newblock \emph{In Proceedings of the International Conference on Learning
  Representations}.

\end{thebibliography}

\clearpage
\newpage
\appendix
\section{Supplement of Experimental Setup}
\label{sec:sup_exp}

\paragraph{Models.}
The CNN model~\cite{kim2014convolutional} contains a single convolutional layer with filter sizes ranging from 3 to 5. The LSTM~\cite{hochreiter1997long} has a single unidirectional hidden layer. We adopt the pretrained BERT-base and DeBERTa-base models from Hugging Face\footnote{\url{https://github.com/huggingface/pytorch-transformers}{}}. We implement the models in PyTorch 1.7.

\paragraph{Datasets.}
 We clean up the text by converting all characters to lowercase, removing extra whitespaces and special characters. We tokenize texts and remove low-frequency words to build vocab. We truncate or pad sentences to the same length for mini-batch during training. Table \ref{tab:preprocess} shows pre-processing details on the datasets.
 
 \begin{table}
 	\centering
 	\begin{tabular}{cccc}
 		\toprule
 		Datasets & $vocab$ & $threshold$ & $length$  \\
 		\midrule
 		SST2 & 13838 & 0 & 50 \\
 		IMDB & 29571 & 5 & 250  \\
 		AG & 21821 & 5 & 50 \\
 		TREC & 8095 & 0 & 15 \\
 		\bottomrule
 	\end{tabular}
 	\caption{Pre-processing details on the datasets. $vocab$: vocab size; $threshold$: low-frequency threshold; $length$: mini-batch sentence length.}
 	\label{tab:preprocess}
 \end{table}

\paragraph{Attack methods.}
We conducted all the adversarial attacks on the TextAttack benchmark \citep{morris2020textattack} with default settings.
\begin{enumerate}
	\item Textfooler \citep{jin2020bert}: Textfooler generates adversarial examples by replacing important words with synonyms from the counter-fitting word embedding space \citep{mrkvsic2016counter}. Part-of-speech (POS) checking and semantic similarity checking via Universal Sentence Encoder (USE) are adopted to select high-quality adversarial examples. 
	\item PWWS \citep{ren2019generating}: PWWS generates adversarial examples by replacing words with their synonyms in WordNet \citep{miller1998wordnet} and replacing named entities (NEs) with similar NEs. 
	\item Gene \citep{alzantot2018generating}: Gene perturbs percentage of words with their nearest neighbors in the GloVe embedding space \citep{pennington-etal-2014-glove} and generates adversarial examples via genetic algorithms.
	\item IGA \citep{wang2019natural}: IGA is an improved genetic based text attack method \citep{alzantot2018generating} which allows to substitute a word with their synonyms in the same position more than once.
	\item PSO \citep{zang-etal-2020-word}: PSO generates adversarial examples by swapping words from HowNet \citep{dong-etal-2010-hownet}.
	\item Clare \citep{li2020contextualized}: Clare is a context-aware adversarial attack method. It applies masks on inputs and plugs in an alternative
	using a pretrained masked language model RoBERTa \citep{liu2019roberta}. 
	\item BAE \citep{garg-ramakrishnan-2020-bae}: BAE replaces and inserts tokens in the original text by masking a portion of the text and leveraging a BERT masked language model to generate alternatives for the masked tokens.
\end{enumerate}

All experiments were performed on a single NVidia GTX 1080 GPU.

\section{Validation Performance}
\label{sec:val_run}
The corresponding validation accuracy for each reported test accuracy is in Table \ref{tab:val-acc}.

\begin{table}
 	\small
	\centering
	\begin{tabular}{p{3.8cm}p{0.6cm}p{0.6cm}P{0.6cm}p{0.6cm}}
		\toprule
		Models & SST2 & IMDB & AG & TREC \\
		\midrule
		LSTM-base & 85.21  & 88.42  & 90.68  & 89.16  \\
		LSTM-adv(Textfooler) & 83.95  & 88.52  & 89.63  & 88.72  \\
		LSTM-adv(PWWS) & 83.49  & 88.56  & 90.97  & 87.39  \\
		LSTM-\ourmethod(Textfooler) & 84.29  & 89.02  & 90.70  & 89.38  \\
		LSTM-\ourmethod(PWWS) & 84.52  & 88.60 & 91.25  & 90.27  \\
		\midrule
		CNN-base & 84.06  & 88.50  & 90.97  & 88.27  \\
		CNN-adv(Textfooler) & 82.22  & 88.94  & 91.07  & 88.05  \\
		CNN-adv(PWWS) & 82.91  & 88.84  & 91.03  & 89.16  \\
		CNN-\ourmethod(Textfooler) & 83.72  & 88.78  & 91.40  & 88.70  \\
		CNN-\ourmethod(PWWS) & 82.22  & 88.90  & 90.93  & 88.05  \\
		\midrule
		BERT-base & 91.63  & 85.52  & 94.63  &  94.25 \\
		BERT-adv(Textfooler) & 91.97  & 87.16  & 90.48  &  94.47 \\
		BERT-adv(PWWS) & 90.67  &  94.02 & 94.52  & 94.47  \\
		BERT-\ourmethod(Textfooler) & 91.40  & 87.32  & 94.80  &  94.91 \\
		BERT-\ourmethod(PWWS) & 91.74  & 87.28  & 94.23  & 94.69 \\
		\midrule
		DeBERTa-base & 93.81  & 88.64  & 94.25  & 94.69  \\
		DeBERTa-adv(Textfooler) & 93.92  & 87.64  & 94.47  & 95.13  \\
		DeBERTa-adv(PWWS) & 94.04  &  89.16 & 94.50  & 92.13  \\
		DeBERTa-\ourmethod(Textfooler) & 93.46  & 89.52  & 94.73  & 94.69  \\
		DeBERTa-\ourmethod(PWWS) & 93.35  & 89.44  & 94.68  & 95.13 \\
		\bottomrule
	\end{tabular}
	\caption{Validation accuracy (\%) for each reported test accuracy.}
	\label{tab:val-acc}
\end{table}

\section{Supplement of Quantitative Evaluations}
\label{sec:sup_quan_eva}

The results of model robustness with respect to LIME interpretations are in Table \ref{tab:lime_inter}, showing the same tendency as those of IG interpretations reported in Table \ref{tab:robustness}.

\begin{table*}[h] 
	\centering
	\begin{tabular}{cccccccccc}
		\toprule
		& & \multicolumn{2}{c}{SST2} & \multicolumn{2}{c}{IMDB} & \multicolumn{2}{c}{AG} & \multicolumn{2}{c}{TREC} \\
		\cmidrule(lr){3-4} \cmidrule(lr){5-6}\cmidrule(lr){7-8} \cmidrule(lr){9-10}
		Attacks & Models & KT & TI & KT & TI & KT & TI & KT & TI \\
		\midrule
		\multirow{9}{*}{Textfooler} & LSTM-base & 0.33 & 0.65 & 0.39 & 0.50 & 0.27 & 0.41 & 0.20 & 0.83 \\
		& LSTM-adv & 0.36 & 0.67 & 0.39 & 0.53 & 0.25 & 0.39 & 0.36 &  0.83 \\
		& LSTM-\ourmethod & \textbf{0.39} & \textbf{0.69} & \textbf{0.41} & \textbf{0.54}  & \textbf{0.41} & \textbf{0.52}  & \textbf{0.45} & \textbf{0.87} \\
		\cmidrule(lr){2-10}
		& CNN-base & 0.33 & 0.67 & 0.44 & 0.51 & 0.25 & 0.37 & 0.27 & 0.83  \\
		& CNN-adv & 0.35 & 0.68 & 0.46 & 0.53 & 0.25 & 0.38 & 0.32 & 0.84 \\
		& CNN-\ourmethod & \textbf{0.39} & \textbf{0.69} & \textbf{0.50} & \textbf{0.56}  & \textbf{0.26} & \textbf{0.40}  & \textbf{0.40} & \textbf{0.87}  \\
		\cmidrule(lr){2-10}
		& BERT-base & 0.14 & 0.55 & 0.13 & 0.20 & 0.07 & 0.23 & 0.08 & 0.79  \\
		& BERT-adv & 0.13 & 0.54 & 0.10 & 0.17 & 0.07 & 0.23 & 0.10 & 0.79 \\
		& BERT-\ourmethod & \textbf{0.21} & \textbf{0.60} & \textbf{0.14} & \textbf{0.21}  & \textbf{0.11} & \textbf{0.25}  & \textbf{0.12} & \textbf{0.81}  \\
		\cmidrule(lr){2-10}
		& DeBERTa-base & 0.21 & 0.55 & 0.17 & 0.25 & 0.08 & 0.23 & 0.13 & 0.81  \\
		& DeBERTa-adv & 0.20 & 0.54 & 0.12 & 0.19 & 0.08 & 0.23 & 0.13 &  0.81 \\
		& DeBERTa-\ourmethod & \textbf{0.22} & \textbf{0.56} & \textbf{0.19} & \textbf{0.26}  & \textbf{0.09} & \textbf{0.24}  & \textbf{0.14} & \textbf{0.82}  \\
		\midrule
		\multirow{9}{*}{PWWS} & LSTM-base & 0.36 & 0.67 & 0.39 & 0.52 & 0.26 & 0.41 & 0.19 & 0.82  \\
		& LSTM-adv & 0.38 & 0.68 & 0.35 & 0.52 & 0.26 & 0.41 & 0.22 & 0.82  \\
		& LSTM-\ourmethod & \textbf{0.42} & \textbf{0.71} & \textbf{0.40} & \textbf{0.55}  & \textbf{0.27} & \textbf{0.43}  & \textbf{0.25} & \textbf{0.83}  \\
		\cmidrule(lr){2-10}
		& CNN-base & 0.37 & 0.68 & 0.43 & 0.52 & 0.26 & 0.39 & 0.19 & 0.83  \\
		& CNN-adv & 0.43 & 0.71 & 0.47 & 0.54 & 0.26 & 0.41  & 0.22 &  0.84 \\
		& CNN-\ourmethod & \textbf{0.47} & \textbf{0.73} & \textbf{0.51} & \textbf{0.55}  & \textbf{0.27} & \textbf{0.42}  & \textbf{0.25} & \textbf{0.85}  \\
		\cmidrule(lr){2-10}
		& BERT-base & 0.14 & 0.55 & 0.13 & 0.21 & 0.07 & 0.22 & 0.08 & 0.79  \\
		& BERT-adv & 0.11 & 0.54 & 0.11 & 0.17 & 0.07 & 0.22 & 0.11 & 0.79  \\
		& BERT-\ourmethod & \textbf{0.25} & \textbf{0.60} & \textbf{0.24} & \textbf{0.31}  & \textbf{0.08} & \textbf{0.24}  & \textbf{0.13} & \textbf{0.80}  \\
		\cmidrule(lr){2-10}
		& DeBERTa-base & 0.22 & 0.55 & 0.16 & 0.23 & 0.08 & 0.23 & 0.12 & 0.80  \\
		& DeBERTa-adv & 0.18 & 0.53 & 0.15 & 0.21 & 0.08 & 0.23 & 0.12 &  0.80 \\
		& DeBERTa-\ourmethod & \textbf{0.23} & \textbf{0.57} & \textbf{0.19} & \textbf{0.25}  & \textbf{0.09} & \textbf{0.24}  & \textbf{0.13} & \textbf{0.81}  \\
		\bottomrule
	\end{tabular}
	\caption{Model robustness to adversarial attacks in terms of LIME interpretations. KT: Kendall’s Tau order rank correlation; TI: top-k intersection ($k=5$).}
	\label{tab:lime_inter}
\end{table*}

\begin{figure*}[h]
	\centering
	\subfigure[LSTM, SST2]{
		\label{fig:tf_ig_lstm_sst2}
		\includegraphics[width=0.2\textwidth]{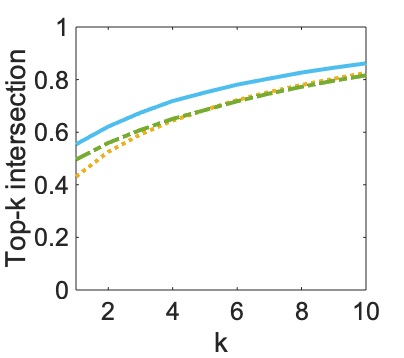}}
	\subfigure[LSTM, IMDB]{
		\label{fig:tf_ig_lstm_imdb}
		\includegraphics[width=0.2\textwidth]{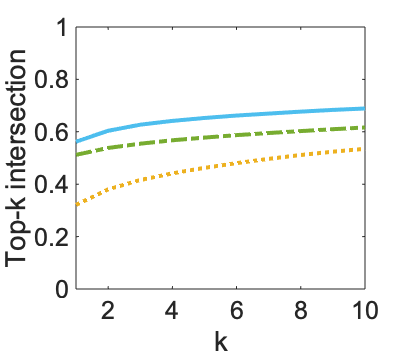}}
	\subfigure[LSTM, AG]{
		\label{fig:tf_ig_lstm_ag}
		\includegraphics[width=0.2\textwidth]{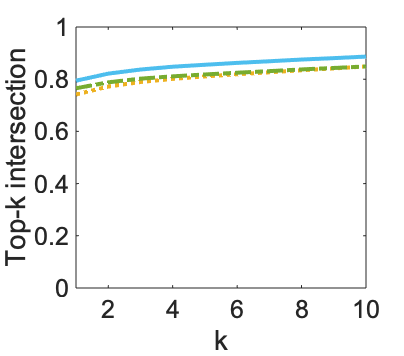}}
	\subfigure[LSTM, TREC]{
		\label{fig:tf_ig_lstm_trec}
		\includegraphics[width=0.2\textwidth]{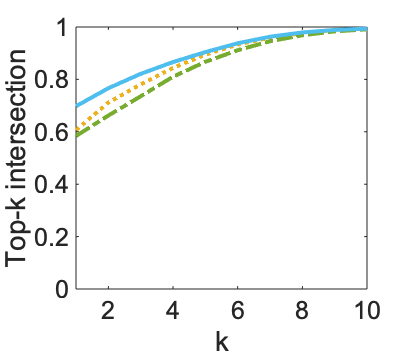}}
	\subfigure[CNN, SST2]{
		\label{fig:tf_ig_cnn_sst2}
		\includegraphics[width=0.2\textwidth]{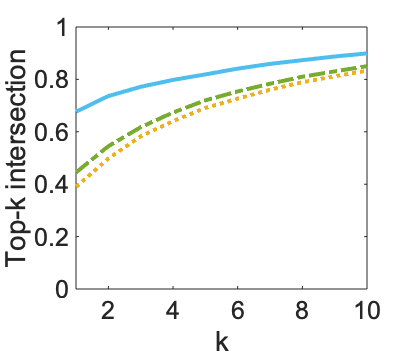}}
	\subfigure[CNN, IMDB]{
		\label{fig:tf_ig_cnn_imdb}
		\includegraphics[width=0.2\textwidth]{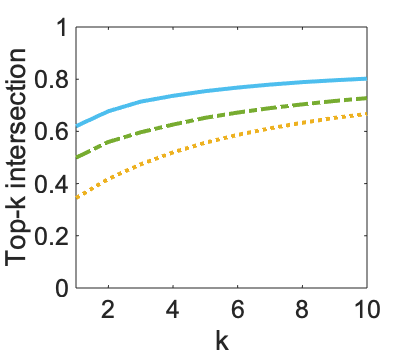}}
	\subfigure[CNN, AG]{
		\label{fig:tf_ig_cnn_ag}
		\includegraphics[width=0.2\textwidth]{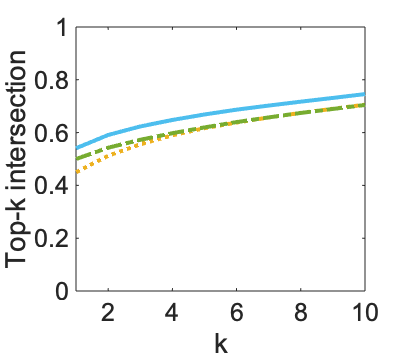}}
	\subfigure[CNN, TREC]{
		\label{fig:tf_ig_cnn_trec}
		\includegraphics[width=0.2\textwidth]{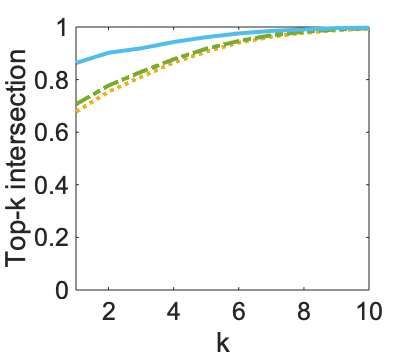}}
	\subfigure[BERT, SST2]{
		\label{fig:tf_ig_bert_sst2}
		\includegraphics[width=0.2\textwidth]{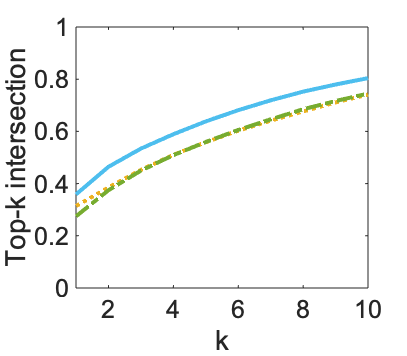}}
	\subfigure[BERT, IMDB]{
		\label{fig:tf_ig_bert_imdb}
		\includegraphics[width=0.2\textwidth]{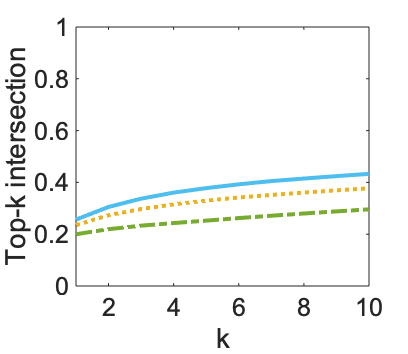}}
	\subfigure[BERT, AG]{
		\label{fig:tf_ig_bert_ag}
		\includegraphics[width=0.2\textwidth]{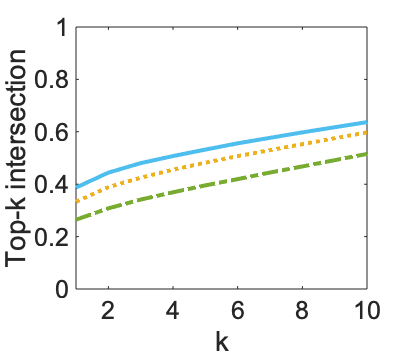}}
	\subfigure[BERT, TREC]{
		\label{fig:tf_ig_bert_trec}
		\includegraphics[width=0.25\textwidth]{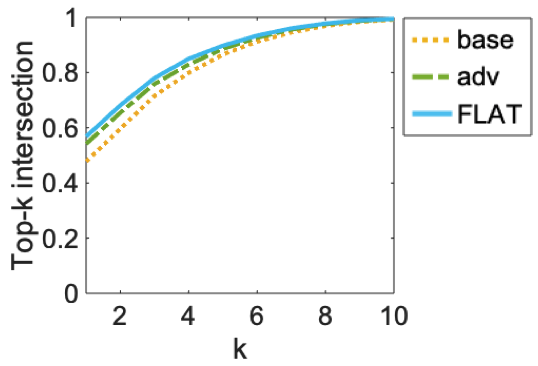}}
	\subfigure[DeBERTa, SST2]{
		\label{fig:tf_ig_deberta_sst2}
		\includegraphics[width=0.2\textwidth]{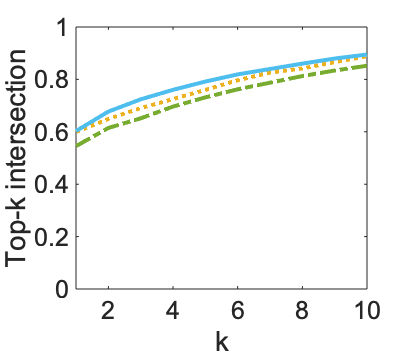}}
	\subfigure[DeBERTa, IMDB]{
		\label{fig:tf_ig_deberta_imdb}
		\includegraphics[width=0.2\textwidth]{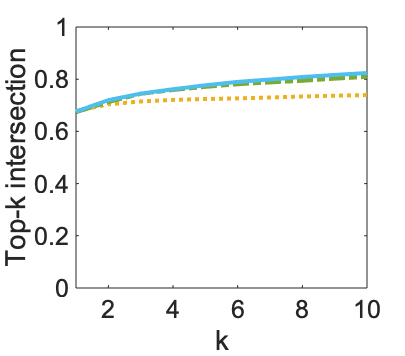}}
	\subfigure[DeBERTa, AG]{
		\label{fig:tf_ig_deberta_ag}
		\includegraphics[width=0.2\textwidth]{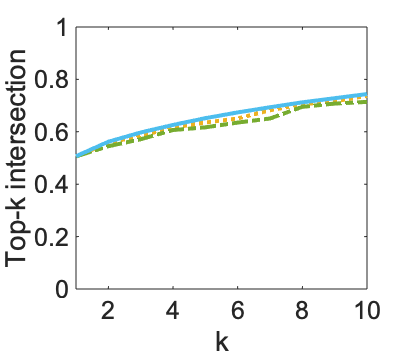}}
	\subfigure[DeBERTa, TREC]{
		\label{fig:tf_ig_deberta_trec}
		\includegraphics[width=0.2\textwidth]{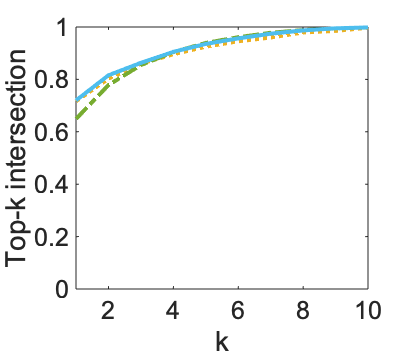}}
	\caption{Top-k intersection of IG interpretations for different models on the four datasets under the Textfooler attack with k increasing from 1 to 10.}
	\label{fig:top_k_ig}
\end{figure*}

The top-k intersections of IG interpretations under the Textfooler attack with k increasing from 1 to 10 are shown in Fig. \ref{fig:top_k_ig}.

\section{Supplement of Discussion}
\label{sec:sup_discussion}

\begin{figure}
	\centering
	\includegraphics[width=0.4\textwidth]{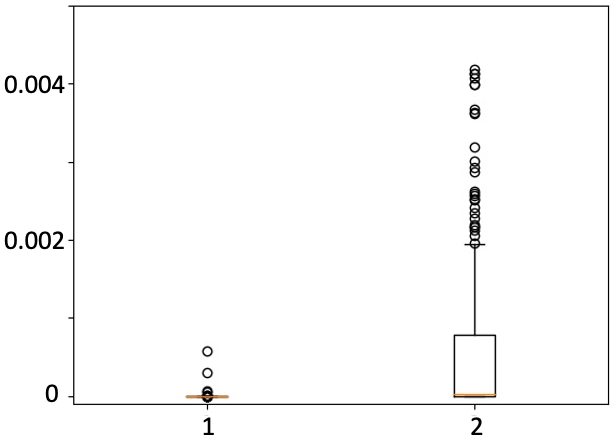}
	\caption{\label{fig:box_plot} Box plot of the L1 norm on the difference between the global importance scores of 1000 randomly sampled words and 10 of their synonyms. 1: $\beta=0.1$, $\gamma=0.001$; 2: $\beta=0.1$, $\gamma=0$.}
\end{figure}

\paragraph{Ablation study.}
To evaluate the effect of the regularization on global importance scores of the replaced words and their substitutions, we compare the LSTM-\ourmethod model trained with parameters $\beta=0.1$, $\gamma=0.001$ and the other one trained with $\beta=0.1$, $\gamma=0$. Fig. \ref{fig:box_plot} shows the box plot of L1 norm on the difference between the learned global importance of 1000 randomly sampled words and 10 of their synonyms in the counter-fitting word embedding space \citep{mrkvsic2016counter}. The results show that the model trained without the regularization ($\beta=0.1$, $\gamma=0$) has larger variance on the global importance scores of synonyms, which makes the model fail to recognize some words and their substitutions as the same important, resulting in relatively low after-attack accuracy.

\paragraph{Correlations between global word importance, word frequency, and word substitution frequency.}

The scatter plots in Fig. \ref{fig:corr} show the correlations between global word importance, word frequency, and word substitution frequency of LSTM-\ourmethod on the SST2 dataset. Any two of the three do not have strong correlations, as their Pearson correlations show in Table \ref{tab:correlation}. Figure \ref{fig:corr} (a) shows that adversarial examples attack the model vulnerability by replacing some words in original input texts, while the replaced words are not based on their frequency. Figure \ref{fig:corr} (b) and (c) show that global word importance scores were learned during training, not trivially based on word frequency or substitution frequency. It is obvious the words that have high substitution frequency in adversarial examples have high importance scores. \ourmethod can also recognize important words with low frequency or even not replaced as they are important for model predictions.

\begin{table}[h] 
	\centering
	\begin{tabular}{ccc}
		\toprule
		Correlation & Coefficient r & P-value \\
		\midrule
		WI - WF & -0.028 & 0.001 \\
		WI - SF & 0.087 & 5.79e-5 \\
		WF - SF & 0.162  & 1.65e-8 \\
		\bottomrule
	\end{tabular}
	\caption{Pearson correlations between global word importance (WI), word frequency (WF), and word substitution frequency (SF).}
	\label{tab:correlation}
\end{table}

\paragraph{A trivial case of \ourmethod.}
\ourmethod automatically learns global word importance and regularizes the importance between word pairs based on the model vulnerability detected by adversarial examples, rather than simply giving the same importance to synonyms. We designed a trivial method for comparison in our pilot experiments. Specifically, we cluster similar words into different groups based on their distance in the counter-fitting word embedding space \citep{mrkvsic2016counter}. Each group of words share the same mask with the value initialized as 1. We train a CNN model with the group masks which are applied on word embeddings on the AG dataset. We test the model robustness with Textfooler attack.

Table \ref{tab:cluster_acc} shows model performance with different predefined group numbers. This trivial method can not help improve model prediction performance on either the standard test set or adversarial examples, even with the number of groups increased. Besides, the embedding space used for clustering words is specified and the number of clusters is predefined, which limit the applicability of this method to defend more complex adversarial attacks. Motivated by these observations, we proposed \ourmethod to automatically learn and adjust word importance for improving model robustness via adversarial training.

\begin{table}[h] 
	\small
	\centering
	\begin{tabular}{ccc}
		\toprule
		Group number & SA & AA \\
		\midrule
		10 & 91.13 & 1.20 \\
		20 & 90.97 & 1.92 \\
		30 & 91.13 & 1.40 \\
		40 & 91.13 & 1.46 \\
		50 & 91.09 & 1.38 \\
		\bottomrule
	\end{tabular}
	\caption{Model performance with different group numbers. SA: standard accuracy (\%); AA: after-attack accuracy (\%)}
	\label{tab:cluster_acc}
\end{table}

\section{Hyperparameters for Reproduction}
We tune hyperparameters manually for each model to achieve the best prediction accuracy on standard test sets. We experiment with different hyperparameters, such as learning rate $lr \in \{1e-5, 1e-3,\cdots, 0.1\}$, clipping norm $clip \in \{1e-3, 1e-2, \cdots, 1, 5, 10\}$, $\beta \in \{0.00001, 0.1, \cdots, 1000\}$, and $\gamma \in \{0.00001, 0.1, \cdots, 1000\}$. All reported results are based on one run for each setting.

\end{document}